\documentclass[letterpaper, 10 pt, conference]{ieeeconf}  % Comment this line out if you need a4paper

\IEEEoverridecommandlockouts                              % This command is only needed if 
\usepackage{graphics} % for pdf, bitmapped graphics files
\usepackage{amsmath}
\usepackage{amssymb}
\usepackage{mathtools}
\usepackage{bm}
\usepackage[T2A,T1]{fontenc}
\usepackage{hyperref}
\usepackage{microtype}
\usepackage{graphicx}
\usepackage{booktabs}
\usepackage{eso-pic} % used by \AddToShipoutPicture
\usepackage{url}
\usepackage{times} % assumes new font selection scheme installed
\usepackage{algorithm} % For algorithms
\usepackage{algpseudocode}
\usepackage{balance}
\usepackage{amssymb}

\def\eg{{\it e.g.}}

\def\ie{{\it i.e.}}

\title{\LARGE \bf
TSPDiffuser: Diffusion Models as Learned Samplers for\\Traveling Salesperson Path Planning Problems
}

\author{Ryo Yonetani$^{1}$% <-this % stops a space
\thanks{$^{1}$Ryo Yonetani is with CyberAgent, Inc., Tokyo, Japan. 
        {\tt\small yonetani\_ryo@cyberagent.co.jp}}%
}

\begin{document}

\maketitle

\thispagestyle{empty}
\pagestyle{empty}

%%%%%%%%%%%%%%%%%%%%%%%%%%%%%%%%%%%%%%%%%%%%%%%%%%%%%%%%%%%%%%%%%%%%%%%%%%%%%%%%
\begin{abstract}
This paper presents TSPDiffuser, a novel data-driven path planner for traveling salesperson path planning problems (TSPPPs) in environments rich with obstacles. Given a set of destinations within obstacle maps, our objective is to efficiently find the shortest possible collision-free path that visits all the destinations. In TSPDiffuser, we train a diffusion model on a large collection of TSPPP instances and their respective solutions to generate plausible paths for unseen problem instances. The model can then be employed as a learned sampler to construct a roadmap that contains potential solutions with a small number of nodes and edges. This approach enables efficient and accurate estimation of travel costs between destinations, effectively addressing the primary computational challenge in solving TSPPPs. Experimental evaluations with diverse synthetic and real-world indoor/outdoor environments demonstrate the effectiveness of TSPDiffuser over existing methods in terms of the trade-off between solution quality and computational time requirements.
\end{abstract}

\section{Introduction}
\label{sec:intro}

Imagine an automated delivery robot tasked with delivering packages to several desks within a cluttered office. To ensure high levels of efficiency and safety, the robot must accomplish two tasks: determine the order of the desks to visit and find collision-free paths between consecutive destinations in obstacle-rich environments. Similar applications may include robotic security patrols~\cite{basilico2022recent} and inventory management~\cite{carreras2013store}, which we envision will be realized in the near future in unconstrained street corners and stores.

We are interested in a composite problem of traveling salesperson problems (TSPs)~\cite{matai2010traveling} and path planning introduced above, which we refer to as \emph{traveling salesperson path planning problems (TSPPPs)}. To solve a TSP for the shortest possible loop that visits all destinations, travel costs must be estimated in advance for all pairs of destinations via path planning. While various implementations of fast TSP solvers are readily available (\eg, Google OR-Tools Routing Library~\cite{ortools_routing}), path planning algorithms require longer computation times for higher solution quality (\ie, success rate and path optimality). This poses the main computational challenge in solving TSPPPs, especially when the obstacle layout in the environment can change regularly.

Machine learning has attracted increasing attention as a promising tool for enabling efficient, high-quality path planning in a data-driven manner. Given a collection of path planning problem instances and their solutions, data-driven path planners learn a neural network to directly generate solution paths~\cite{qureshi2019motion,nasiriany2019planning,chen2021decision,janner2021offline,janner2022planning,liang2023adaptdiffuser} or roadmaps/trees on which solution paths can be found efficiently via sampling-based path planning~\cite{ichter2018learning,ichter2019robot,ichter2020learned,faust2018prm,chen2019learning}. However, most existing data-driven planners are specialized for a single-query setup, \ie, require the exploration of the same environment for every pair of starts and goals. This makes these planners an inefficient choice for TSPPPs that need to execute path planning as many times as the square of the number of destinations for every obstacle layout.

\begin{figure}[t]
    \centering
    \includegraphics[width=\linewidth]{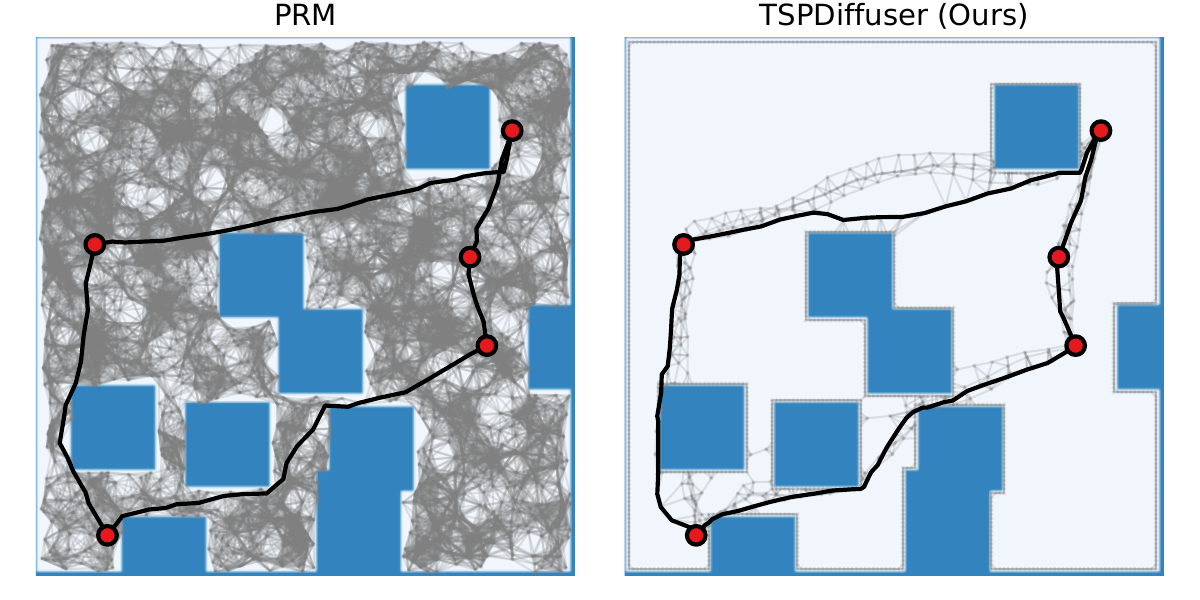}
    \caption{\textbf{Solving TSPPPs with TSPDiffuser (right).} We leverage a diffusion model as a learned sampler to construct a roadmap for traveling salesperson path planning problems (TSPPPs). Our approach significantly improves the balance between solution quality and efficiency compared to existing sampling-based path planners such as probabilistic roadmaps (PRM; left)}
    \label{fig:teaser}
\end{figure}

In this work, we propose \emph{TSPDiffuser}, a novel data-driven path planner designed to solve TSPPPs. At its core, we learn a diffusion model~\cite{ho2020denoising,yang2023diffusion} to generate plausible loop paths from obstacle and destination maps. Rather than attempting to generate a single solution path directly from a learned model, TSPDiffuser samples multiple paths to construct a roadmap for a higher success rate and path optimality as depicted in Fig.~\ref{fig:teaser}. Because this roadmap highly likely contains potential solutions with a small number of nodes and edges, it alone can be used to efficiently estimate travel costs for all destination pairs. In this way, TSPDiffuser aims to achieve both high solution quality and computational efficiency, addressing the main challenge for TSPPPs.

We extensively evaluate our approach using three synthetic environments with diverse obstacle layouts, as well as 30 outdoor and 15 indoor environments with more complex layouts of obstacles scanned from real-world buildings, walls, or furniture~\cite{sturtevant2012benchmarks,dobrevski2020adaptive,xia2018gibson}.
Experimental results demonstrate that TSPDiffuser significantly improves the trade-off between solution quality and computation time over existing methods~\cite{janner2022planning,ichter2020learned,karaman2011sampling}.

\begin{figure*}[t]
    \centering
    \includegraphics[width=\linewidth]{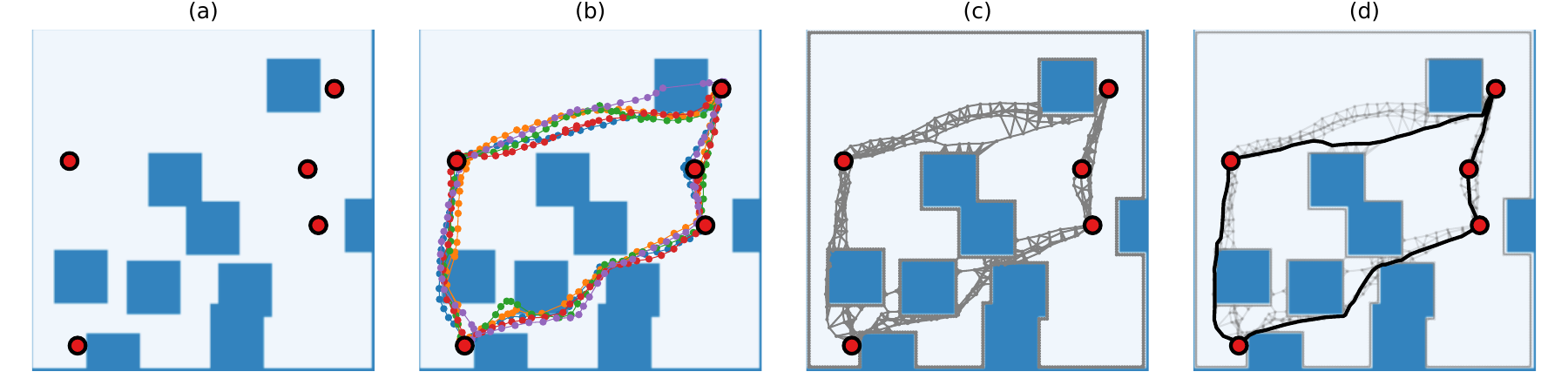}
    \caption{\textbf{Solution Overview.} (a)~Problem instance. (b)~Generating multiple paths (each shown in different colors) using a learned diffusion model. (c)~Connecting paths to construct a roadmap. (d)~Perform path planning and solving TSP on the roadmap to obtain a solution path.}
    \label{fig:graph}
\end{figure*}
\section{Related Work}
\label{sec:related_work}

\paragraph{Path Planning and Data-Driven Planners}
Path planning is a classical problem in AI and robotics, where the goal is to find the lowest-cost feasible path between two states (\eg, positions, poses). Particularly in continuous or high-dimensional state spaces, sampling-based approaches such as probabilistic roadmaps (PRM)~\cite{kavraki1996probabilistic}, rapidly-exploring random trees (RRT)~\cite{lavalle1998rapidly}, and their asymptotically optimal versions~\cite{karaman2011sampling} are often used. Recent work aims to leverage machine learning techniques to develop data-driven planners in various ways: learning neural networks that can directly generate paths~\cite{qureshi2019motion}, learning sampling distributions for constructing roadmaps~\cite{ichter2018learning,ichter2020learned,faust2018prm}, or learning policies and/or value functions for tree-based search~\cite{ichter2019robot,chen2019learning}. Machine learning modules in such planners are often integrated into classical planning frameworks to ensure high success rates, thus inheriting the trade-off between planning quality and efficiency; the more search costs we spend, the more likely we are to obtain a feasible and optimal solution if one exists. Moreover, most of the existing data-driven planners except~\cite{ichter2020learned} consider a single-query setup and need to explore every environment for every pair of sources and destinations.

\paragraph{Traveling Salesperson Problem (TSP)}
TSP is a combinatorial optimization problem that aims to find the shortest possible route visiting all given destinations exactly once before returning to the origin, given a set of destinations and pairwise travel costs among them. TSP has a long history in theoretical computer science and operations research and has been extended to robotics tasks such as order picking in warehouses and vehicle routing~\cite{matai2010traveling}. Recent work has attempted to extend machine learning techniques to solve Euclidean TSPs~\cite{graikos2022diffusion,xin2021neurolkh,zhang2021solving,jin2023pointerformer}. Various other forms of TSPs have also been studied with practical assumptions on agents (\eg, vehicles, unmanned aerial vehicles, or teams of such moving agents), including Dubins TSP~\cite{savla2005point}, stochastic TSP~\cite{adler2016stochastic}, variable-speed TSP~\cite{kuvcerova2021variable}, trajectory-based TSP~\cite{meyer2021trajectory}, and vehicle routing problems~\cite{braekers2016vehicle}. The proposed TSPPP can be viewed as a specific case of \emph{metric} TSPs, where travel costs between destinations are given by the length of shortest possible collision-free paths and thus only satisfy the triangle inequality. Studies in this direction have been relatively limited~\cite{faigl2011application, babel2017curvature}. No prior work has considered its path planning aspect, \ie, \emph{how to accurately and efficiently evaluate travel costs in obstacle-rich environments}\footnote{In parallel to this work, \cite{huang2023pke} has recently studied multi-goal path finding, a problem setting very close to TSPPPs.}.

\paragraph{Diffusion Models}
Diffusion models belong to a family of probabilistic generative models, which are capable of generating high-quality data through an iterative denoising process. This is achieved by progressively injecting noise to disrupt input data, and learning the reverse process that gradually reconstructs the original data back from complete noise (see \cite{yang2023diffusion} for an introductory survey.) Diffusion models have led to significant advances especially in image generation~\cite{rombach2022high,saharia2022photorealistic}, and have also begun to be used in robot learning for modeling environmental dynamics~\cite{janner2022planning,liang2023adaptdiffuser}.

\begin{figure*}[t]
    \centering
    \includegraphics[width=\linewidth]{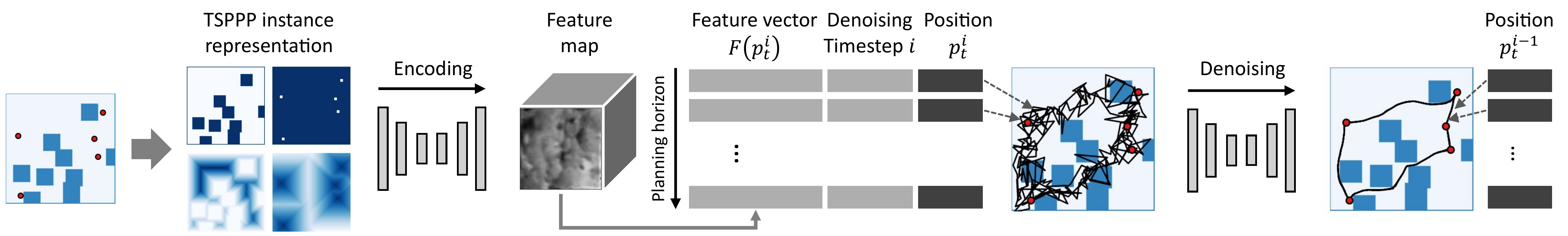}
    \caption{\textbf{Model Architecture.} We represent a TSPPP instance using binary and distance images of obstacles and destinations, which are stacked and encoded into a unified feature map. Noisy position data, concatenated with denoising timestep information and the feature vector extracted from the feature map, are fed into a diffusion model to reconstruct the original clean path.}
    \label{fig:arch}
\end{figure*}

\section{Preliminaries}
\label{sec:preliminaries}

\subsection{Problem Definition}
\label{subsec:problem_def}
Fig.~\ref{fig:graph} provides the overview of our work. An instance for TSPPPs is represented by an environmental map with obstacles and a set of destinations (Fig.~\ref{fig:graph}a). The solution is expected to be a path that visits all the destinations while avoiding obstacles and returning to the starting point (Fig.~\ref{fig:graph}d). The quality of solution paths is measured by feasibility (\ie, if the path is collision-free) and optimality (\ie, how close the path is to the shortest one.) The performance of planners is evaluated by the trade-off between solution quality and computational efficiency, or in practice, the computation time required to derive solutions.

Our primary objective is to improve this trade-off for solving TSPPPs. Particularly in mobile robotics applications, scalability with respect to the number of destinations can inherently be limited due to battery constraints. Instead, quick and high-quality path planning becomes more critical to increasing overall task throughput, because it must be used thoroughly to estimate travel costs between destinations for every change in the obstacle layout in the environment.

Concretely, let $\Omega = [-1, 1]^2$ be a continuous state space, where the free (\ie, non-obstacle) space is given by $\Omega_\mathrm{free}\subset\Omega$ and the position of a robotic agent, as well as its destinations, are represented by a 2D real-valued vector $p\in\Omega_\mathrm{free}$. We represent the free space by a roadmap $G=(V, E)$, where nodes $V\subset \Omega_\mathrm{free}$ represent positions and edges $E \subset V\times V$ connect the nodes with their neighbors if there are no obstacles between them. Valid edges have the Euclidean distance between the positions as edge costs. Let $\Omega_\mathrm{dst}\subset \Omega_\mathrm{free}$ be a set of destinations to visit, which we assume to be included in the vertex set as well, \ie, $\Omega_\mathrm{dst}\subset V$. With a graph search algorithm, we can find a shortest path and its cost between destinations on the constructed roadmap.

\subsection{Denoising Diffusion Probabilistic Models (DDPMs)}
\label{subsec:diffusion_models}
We propose to leverage diffusion models for constructing a roadmap with a small number of significant nodes and edges. We will particularly adopt denoising diffusion probabilistic models (DDPMs)~\cite{ho2020denoising} that model perturbation and denoising processes by Markov chains, although any diffusion model could be used in principle. Let $\tau^0$ be a clean, original data sample and $\tau^i$ be its perturbed version by progressively injecting noise $i$ times ($1\leq i \leq I$). In DDPMs, the reverse transition from $\tau^{i}$ to $\tau^{i-1}$ is typically given by the Gaussian with learnable parameters $\theta$~\cite{janner2022planning}:
\begin{align}
    p_\theta(\tau^{i-1} \mid \tau^{i}) = \mathcal{N}(\tau^{i-1}\mid\mu_\theta(\tau^{i}, i), \Sigma^{i}),
    \label{eq:denoise}
\end{align}
where $\Sigma^i$ is a covariance matrix. The function $\mu_\theta(\tau^i, i)$ is parameterized by noise estimator $\varepsilon_\theta$:
\begin{align}
    \mu_\theta(\tau^{i}, i) = \frac{1}{\sqrt{\alpha^i}}\left(\tau^i - \frac{\beta^i}{\sqrt{1-\bar{\alpha}^i}}\varepsilon_\theta(\tau^i, i)\right),
    \label{eq:noise_estimator}
\end{align}
where $\alpha^i$, $\bar{\alpha}^i$ and $\beta^i$ are inter-dependent hyper-parameters chosen appropriately. The function $\varepsilon_\theta(\tau^i, i)$ is given by neural networks such as U-Net~\cite{ronneberger2015u}, and is learned from training data and their perturbed versions (see \cite{ho2020denoising} for the complete introduction.)

Learned DDPMs can generate clean data from complete noise $\mathcal{N}(\mathbf{0}, \mathbf{I})$ by repeatedly applying the denoising kernel in Eq.~(\ref{eq:denoise}). Particularly for trajectory generation~\cite{janner2022planning,liang2023adaptdiffuser}, \emph{reward (value)-guided sampling} encourages generated samples to meet an objective represented by a differentiable reward (or value) function $\mathcal{J}(\tau)$. Specifically, for sample $\tau^i$, the denoising step is extended as follows~\cite{janner2022planning}:
\begin{align}
\tau^{i-1}\sim\mathcal{N}(\mu + \alpha\Sigma\nabla\mathcal{J}(\mu), \Sigma^i),
\label{eq:sampling}
\end{align}
where $\mu=\mu_\theta(\tau^i, i)$ and $\alpha$ is a scale parameter.
\section{TSPDiffuser}
\label{sec:tsp_diffuser}
In TSPDiffuser, we extend a diffusion model to generate plausible loop paths for a given TSPPP instance (Fig.~\ref{fig:graph}b). The generated paths are interconnected to form a roadmap with a small number of vital nodes and edges around potential solutions (Fig.~\ref{fig:graph}c). This efficient construction of a high-quality roadmap contributes to the overall throughput improvement. By carrying out path planning and solving the TSP on this roadmap, we derive the solution path (Fig.~\ref{fig:graph}d).

\subsection{Model Architecture}
\label{subsec:model_arch}
Fig.~\ref{fig:arch} illustrates the proposed model architecture. We define a sample $\tau^i$ by a path, \ie, a sequence of positions $\tau^i=(p^i_0,\dots,p^i_T),\;p^i_t\in\Omega$, that tours all destinations while avoiding obstacles. Because input instances provide the layout of obstacles and destinations, we propose to feed this information into the noise estimation network $\varepsilon_\theta$, effectively conditioning its generation results.

Specifically, we represent obstacles and destinations by binary images and their distance images obtained via distance transform. These images are stacked and encoded into a $d$-dimensional unified feature map $F$ with fully-convolutional networks. The feature vector extracted at location $p$ from the feature map, denoted as $F(p)\in\mathbb{R}^d$, describes a local context about whether there are obstacles and destinations nearby. Our noise estimator takes the form of $\varepsilon_\theta(\tau^i, i, F(\tau^i))$, where $F(\tau^i)=(F(p^i_1),\dots,F(p^i_T))$ is a sequence of feature vectors extracted along $\tau^i=(p^i_1,\dots,p^i_T)$. In practical implementations, we concatenate the position $p^i_t$, the denoising timestep $i$ encoded into a vector via Fourier transform, and the feature vector $F(p^i_t)$ to form the input to the noise estimator.

\subsection{Constructing Roadmaps using Learned Diffusion Models}
\label{subsec:algo}
Despite their high generation capability, we observe two limitations for diffusion models when used in path planning.  Firstly, there is an inevitable trade-off between the quality of generated samples and the required computation time. Although higher throughput with limited computational resources is desirable for practical robotic applications, efficient sampling often sacrifices the sample quality. Secondly, even with reward-guided sampling, there is no guarantee that generated paths satisfy constraints imposed in TSPPPs, \ie, visiting all destinations in a single loop, while avoiding collisions with any single obstacle. This makes it hard to solve TSPPPs by diffusion models alone.

Addressing these limitations, we propose to leverage a diffusion model as a learned sampler for constructing a roadmap $G=(V, E)$. We sample multiple loop paths from a learned diffusion model with relatively low timestep granularity (\eg, $I=5$, as shown in Fig.~\ref{fig:graph}b), and interconnect these paths and destinations to form a roadmap (Fig.~\ref{fig:graph}c). The resulting roadmap contains potential solution paths with a small number of important nodes and edges. This approach is more accurate than attempting to generate a single high-quality path that satisfies the TSPPP constraints, and is more efficient than building a conventional roadmap that covers the entire free space in the environment.

Alg.~\ref{alg:roadmap} describes the proposed algorithm. Let $\{\tau_1,\dots,\tau_M\}$ be a set of paths generated from a learned diffusion model, where $\tau_m=(p_{m, 1},\dots,p_{m, T_m})$ is a sequence of positions. All the positions contained in the generated paths are regarded as nodes $V$ (L3). With a validity check function $\texttt{isvalid}(\cdot, \cdot;\Omega_\mathrm{free})$ that returns true if there are no obstacles between a pair of positions, we first connect nodes along each generated path to define edges $E$ (L4-L7). After adding destinations $\Omega_\mathrm{dst}$ and nodes around boundaries of the obstacles collided with generated paths, $\{b_1,\dots,b_N\}$, to $V$ (L9), we further span edges between each node and its $K$ nearest neighbors to $E$ (L10-L15).

\begin{algorithm}[t]
\caption{Construct Roadmaps with Diffusion Models}
\label{alg:roadmap}
\begin{algorithmic}[1]
\Require{A collection of paths $\{\tau_1,\dots,\tau_M\}$, $\tau_m=(p_{m, 1},\dots,p_{m, T_m})$, generated from a learned diffusion model; a set of destinations $\Omega_\mathrm{dst}$; validity check function $\texttt{isvalid}(\cdot,\cdot;\Omega_\mathrm{free})$}
\Ensure{Roadmap $G=(V, E)$}
\State $V, E\gets \{\},\{\}$
\For{$m=1,\dots,M$} \Comment{Connect nodes along paths}
    \State $V\gets V\cup \{p_{m, t} \mid p_{m, t} \in \tau_m\}$
    \For{$t\in 1,\dots, T_m-1$}
        % \State $v\gets p_{m, t},\; w\gets p_{m, t+1}$
        \If{$\texttt{isvalid}(p_{m, t}, p_{m, t+1};\Omega_\mathrm{free})$}
        $E = E \cup \{(v, w)\}$
        \EndIf
    \EndFor
\EndFor
\State $V\gets V \cup \Omega_\mathrm{dst} \cup \{b_1,\dots,b_N\}$ \Comment{Add destinations and $N$ boundary nodes for the obstacles collided with generated paths}
\For{$v\in V$} \Comment{Connect nodes with their neighbors}
    \For{$w \in \texttt{K-NN}(v)$} \Comment{$K$ neighbors around $v$}
        \If{$\texttt{isvalid}(v, w;\Omega_\mathrm{free})$}
        $E = E \cup \{(v, w)\}$
        \EndIf
    \EndFor
\EndFor
\end{algorithmic}
\end{algorithm}

\subsection{Solving TSPPP}
\label{subsec:solving_tsp}
On the constructed roadmap, we perform path planning to calculate the shortest-path distance as a travel cost between pairs of destinations. We then solve a TSP using the calculated travel costs. This gives us the order of destinations to visit with the total path cost minimized among all possible tours. Finally, we stitch the shortest paths between destinations in the determined order to generate the solution for TSPPPs.

\begin{figure*}[t]
    \centering
    \includegraphics[width=.96\linewidth]{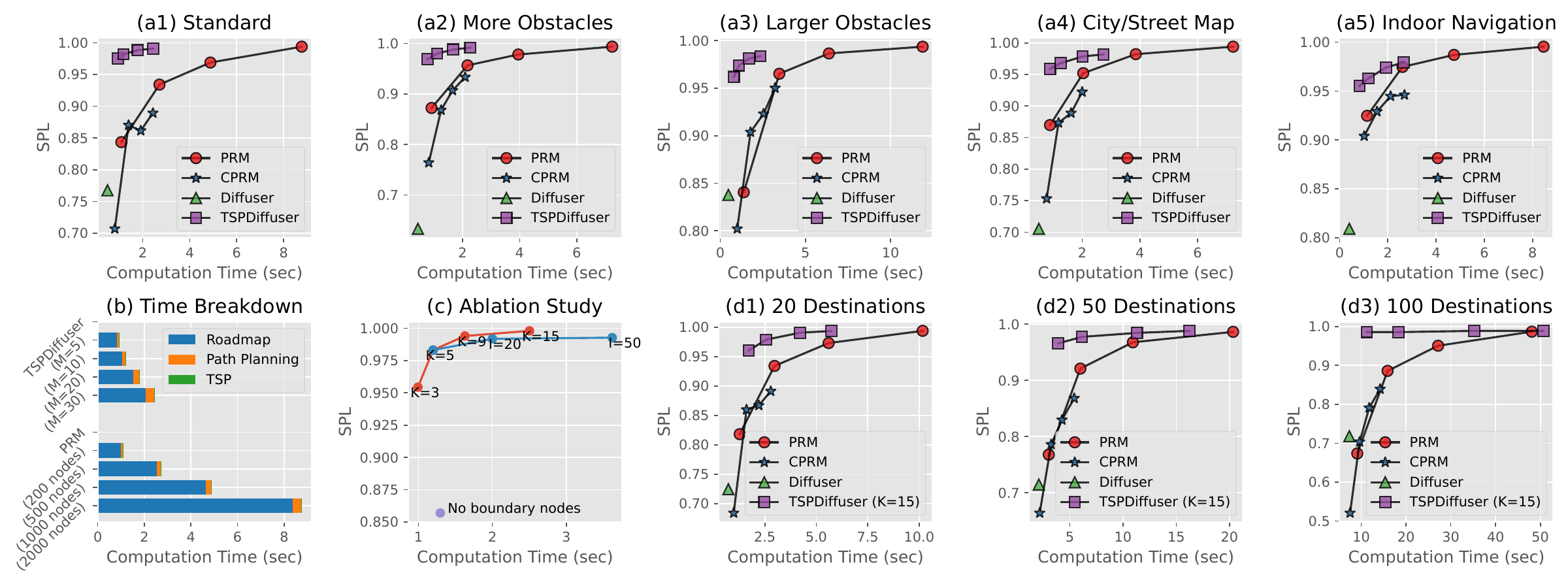}
    \caption{\textbf{Quantitative Results.} (a1-a5, d1-d3)~SPL scores and required computation times. (b)~Specific time breakdown for TSPDiffuser and PRM. (c)~Ablation study.}
    \label{fig:quantitative_result}
\end{figure*}

\begin{figure*}[t]
    \centering
    \includegraphics[width=0.48\linewidth]{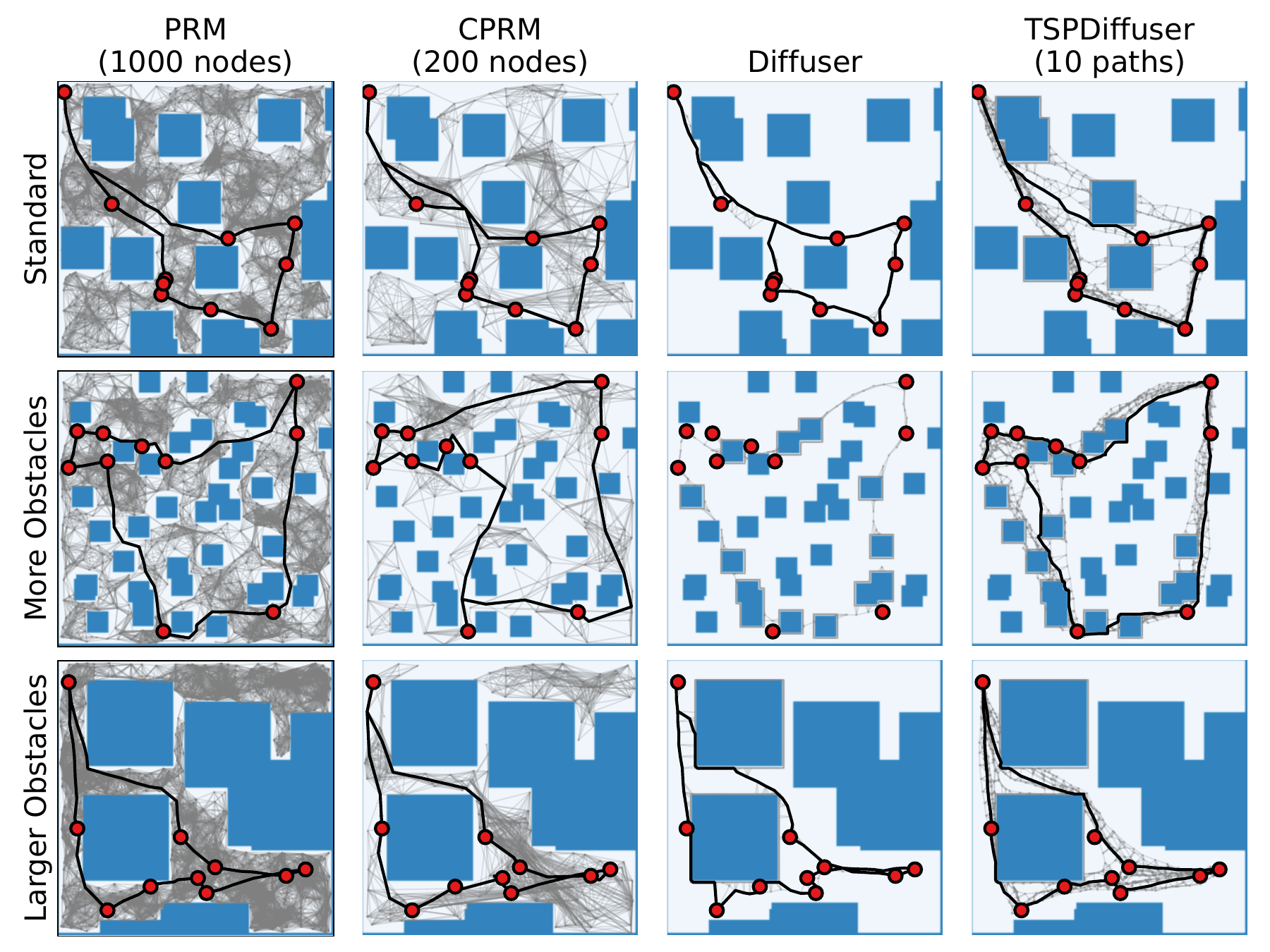}
    \includegraphics[width=0.48\linewidth]{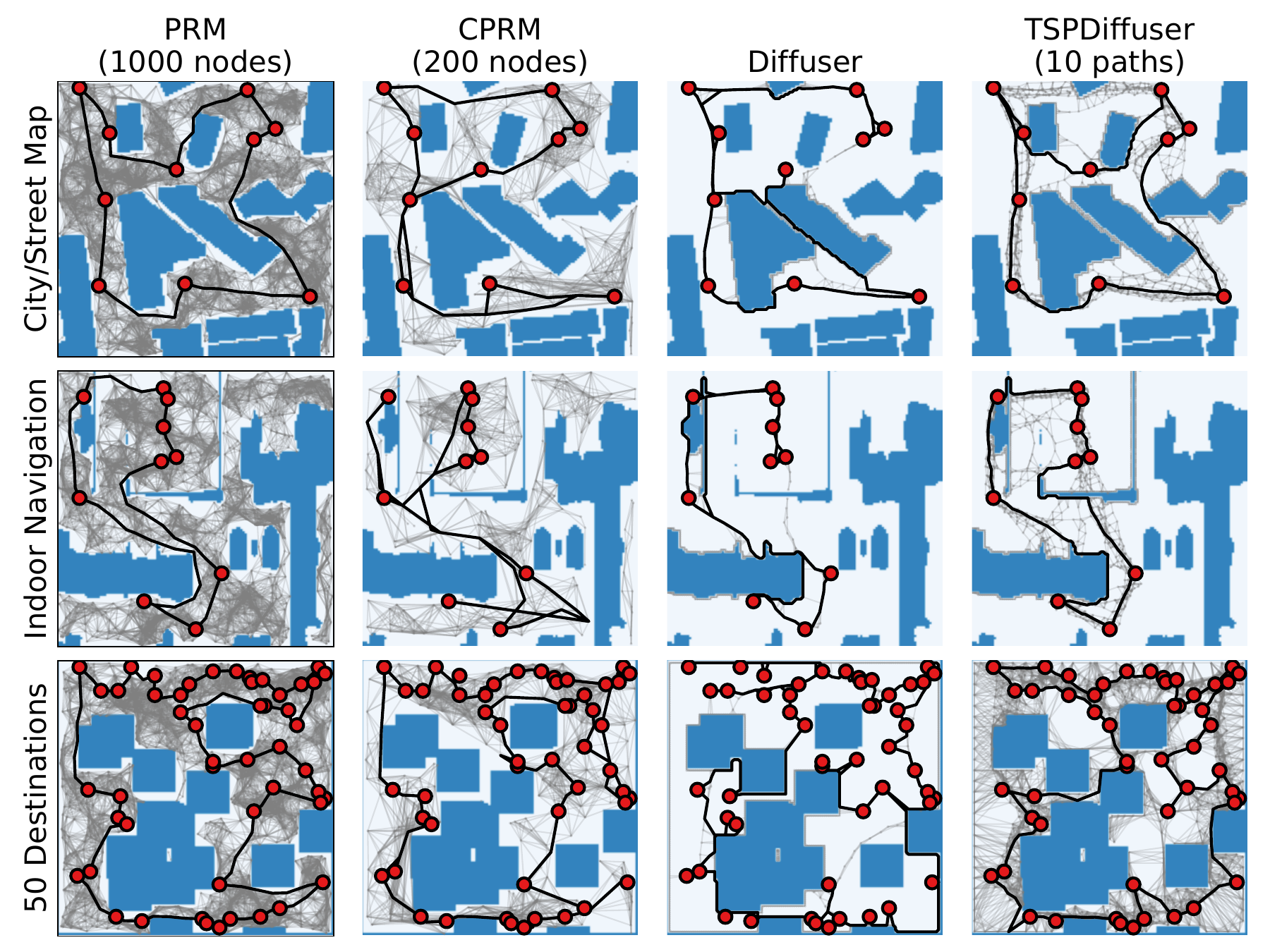}
    \caption{\textbf{Qualitative Results.} Roadmaps and solution paths are visualized with gray and black lines (if solved). Obstacle regions are colored in blue, and destinations are marked with red circles.}
    \label{fig:qualitative_result}
\end{figure*}

\section{Experiments}
\label{sec:experiments}
\subsection{Planning Environments}
We extensively evaluate TSPDiffuser using the following synthetic and real-world indoor/outdoor navigation environments, each with various layouts of obstacles and destinations on 128x128 sized maps:
\begin{itemize}
\item \textbf{Synthetic}: Three synthetic environments with different obstacle layouts; \textbf{Standard} with twenty random 20x20 square-shaped obstacles, \textbf{More Obstacles} with forty random 10x10 square-shaped obstacles, and \textbf{Larger Obstacles} with ten random 40x40 square-shaped obstacles.
\item \textbf{City/Street Map}~\cite{sturtevant2012benchmarks}: Random areas cropped from 30 city/street maps such as Berlin, London, and New York.
\item \textbf{Indoor Navigation Dataset}~\cite{dobrevski2020adaptive}: Random areas cropped from 15 indoor office/living environments for mobile robot navigation tasks, which was generated from Gibson Database~\cite{xia2018gibson}.
\end{itemize}
Note that obstacles in the City/Street Map and Indoor Navigation Dataset are much more complex in their shapes and layouts and faithful to real-world settings, as they are the scans of real-world buildings, walls, or furniture.
For each environment, we created 100 instances by randomizing the locations of obstacles for the Synthetic environments and cropped areas for City/Street Map and Indoor Navigation Dataset. For each instance, ten random destinations were sampled from the largest non-obstacle area. The ground-truth, \ie, (pseudo-) optimal, path for each instance was obtained by computing travel costs among destinations using PRM~\cite{karaman2011sampling} with 3,000 nodes and solving the TSP using \texttt{or-tools}~\cite{ortools_routing}.

In addition, we generated approximately 100,000 instances of the Standard environment as a training dataset for TSPDiffuser and other data-driven baselines presented later. This verifies the generalization ability of these planners from synthetic to real-world environments (\ie, sim-to-real transfer ability). Ground-truth paths in the training dataset were repeated until the total number of temporal steps exceeded 256, and then trimmed to have a fixed step of 256. This process allowed us to easily stack multiple ground-truth paths to form a mini-batch during training.

\subsection{Evaluation Setups}
\paragraph{Evaluation Metric}
For each environment, we calculated the \emph{success rate weighted by path length (SPL)}~\cite{anderson2018evaluation} for 100 instances: $\frac{1}{100}\sum_{i=1}^{100} S_i\frac{\hat{l}_i}{\max (l_i, \hat{l}_i)}$. Here, $S_i$ is the binary indicator that takes $S_i=1$ if the method found a solution path for the $i$-th instance and $S_i=0$ otherwise, and $l_i, \hat{l}_i$ are the lengths of the solution and ground-truth paths. As such, SPL is used to evaluate both success rate and path optimality in a single metric. We evaluate the trend of SPL scores while changing hyper-parameters of a method that affect the trade-off between the solution quality and required computation times, such as the number of nodes for PRMs and that of generated paths $M$ for TSPDiffuser. All the evaluation procedures were run on the Intel\textregistered{} Xeon\textregistered{} CPU @ 2.20GHz.

\paragraph{Baselines}
Due to the absence of prior work that solves the same problem,\footnote{Note that existing learning-based TSP solvers~\cite{graikos2022diffusion,xin2021neurolkh,zhang2021solving,jin2023pointerformer} leverage machine learning to \emph{solve TSPs while assuming that pairwise travel costs are already given}, thus cannot be compared to the proposed method.} we have extended existing path planning algorithms for baseline methods. All the baselines below are able to construct a roadmap to evaluate travel costs for all combinations of destinations (\ie, support multi-query setups), which is then followed by the same solution procedure shown in Sec.~\ref{subsec:solving_tsp}.
\begin{itemize}
    \item \textbf{Probabilistic Roadmap (PRM)}~\cite{karaman2011sampling}: A classical sampling-based path planning algorithm that is also used to generate ground-truth paths. Roadmaps were constructed by sampling $M_\mathrm{prm}\in\{200, 500, 1000, 2000\}$ nodes uniformly from non-obstacle regions, and then connecting the nodes based on $r$-neighbor search, where $r$ was determined adaptively based on \cite{karaman2011sampling}.
    \item \textbf{Critical PRM (CPRM)}~\cite{ichter2020learned}: A data-driven path planner that learns to detect `critical' regions in environments and prioritizes sampling from those regions for constructing a roadmap. In this baseline, we directly regard regions around ground-truth paths as critical, and learn a U-Net to predict those regions. Sampled nodes are connected based on $r$-neighbors, where $r$ was determined in the same way as that of PRM.\footnote{Although the original CPRM connects non-critical nodes to all critical nodes, we employed $r$-neighbors because such full connections required a long time despite limited solution quality in our problem setup.} The number of nodes was selected among $M_\mathrm{cprm}\in\{100, 200, 300, 400\}$ to see how SPL scores and required times can change.
    \item \textbf{Diffuser}~\cite{janner2022planning}: A state-of-the-art trajectory generation method using diffusion models. For a fair comparison, we employed the same diffusion model architecture as that of TSPDiffuser in Sec.~\ref{subsec:model_arch}. A single loop path is generated from a learned model and then connected with destinations and boundary nodes to construct a roadmap. This baseline can be viewed as a degraded version of TSPDiffuser where the number of paths is $M=1$.
\end{itemize}

\paragraph{Implementation Details}
For fairness in evaluation, all methods are implemented on a single codebase written in python, with the same function shared for edge validity check (\ie, $\texttt{isvalid}(\cdot, \cdot;\Omega_\mathrm{free})$ in Sec.~\ref{subsec:algo}.) TSPDiffuser consists of the problem instance encoder network and the noise estimation network. As the instance encoder, we used a U-Net~\cite{ronneberger2015u} with four encoding and decoding layers from the VGG backbone~\cite{simonyan2014very}, which outputs $d=256$-dimensional feature maps. The noise estimation network was a 1D U-Net with residual blocks and self-attention modules, where the input is the sequence of concatenations of positions, denoising timesteps encoded into 16-dimensional real-valued vectors via Fourier transform, and the 256-dimensional feature vectors extracted from the feature maps. These two modules were trained end-to-end with the standard DDPM training pipeline~\cite{ho2020denoising} for 300 epochs with a batch size of 32, using the AdamW optimizer with the cosine annealing learning rate scheduler. The granularity of denoising timesteps, \ie, $I$ in Sec.~\ref{subsec:diffusion_models}, was set to 500 for training, and 5 for sampling. For roadmap construction, the number of sampled paths $M$ were set to either one of $M\in \{5, 10, 20, 30\}$, where larger $M$ will result in higher SPL scores with more computation times. The reward guidance, \ie, $\nabla\mathcal{J}(\tau)$ in Eq.~(\ref{eq:sampling}), is given by the sum of distance images from obstacles and destinations, which guides output positions to avoid obstacles and approach destinations. The number of neighbors considered when spanning edges among paths, \ie, $K$ in Alg.~\ref{alg:roadmap}, were set to $K=5$ unless specified otherwise. The number of boundary nodes, \ie, $N$ in Alg.~\ref{alg:roadmap}, were set adaptively by detecting boundary pixels of obstacles.

\subsection{Results}
\paragraph{Quantitative Results} Fig.~\ref{fig:quantitative_result}a presents the relationships between SPL scores and computation times. Despite being trained on synthetic data, TSPDiffuser generalizes to real-world environments (a4 and a5) without additional training or parameter tuning, demonstrating its transferability to real-robot applications. It achieves high SPL scores with processing speeds that are almost 2x to 4x faster compared to the other baselines, even including model loading and running times on CPUs. This result shows that our approach is advantageous, especially in situations where the obstacle layout in the environment can change regularly. Diffuser operates faster than TSPDiffuser, but its SPL score is significantly lower because a single sampling by diffusion models does not necessarily guarantee feasible or near-optimal paths. CPRM achieved comparable performance with much fewer nodes compared to PRM. Fig.~\ref{fig:quantitative_result}b elaborates on the computation times of TSPDiffuser and PRM. The construction of the roadmap is the main computational bottleneck. TSPDiffuser effectively tackles this problem by limiting the nodes and edges to be sampled around potential solution paths.

\paragraph{Ablation Study} We evaluated variants of TSPDiffuser with different hyper-parameters: (1) the number of $K$-neighbors to connect edges from $K\in\{3, 5, 9, 15\}$, (2) the granularity of denoising timesteps from $I\in\{5, 20, 50, 100\}$, and (3) whether or not nodes along obstacle boundaries are added to roadmaps, to see how these choices impact SPL scores and required times. As summarized in Fig.~\ref{fig:quantitative_result}c, adding boundary nodes is crucial. The number of $K$-neighbors is also important to ensure high SPL scores. Despite increased computation times, no significant improvements were confirmed for higher timestep granularity $I\geq 20$.

\paragraph{Results for More Destinations} We further investigated the scalability of TSPDiffuser with respect to the number of destinations: 20, 50, and 100, on the Standard environment. Based on the ablation study, we evaluated TSPDiffuser with $K=15$ to pursue more optimal solutions at the cost of required computation times. As summarized in Fig.~\ref{fig:quantitative_result}d, TSPDiffuser ($K=15$) consistently outperforms the other baselines. It can find solutions with an SPL higher than $0.98$ in around 10 seconds for instances with 100 destinations, whereas PRM requires nearly 50 seconds to achieve the same level of performance. Note that as the number of destinations significantly increases, paths between them are more likely to be simple straight lines, where a more efficient edge validation method such as visibility roadmaps~\cite{simeon2000visibility} would be effective.

\paragraph{Qualitative Results} Fig.~\ref{fig:qualitative_result} visualizes typical solution paths and roadmaps obtained by each method. Compared to PRM, which uniformly samples nodes, data-driven planners such as CPRM, Diffuser, and TSPDiffuser concentrate more nodes around ground-truth paths. Diffuser sometimes failed to find feasible paths connecting all destinations without collisions. TSPDiffuser can find collision-free paths more reliably by generating diverse paths to form a roadmap.

\subsection{Limitations and Possible Extensions}
TSPDiffuser currently supports only 2D state space. Extending our approach to higher dimensional state spaces is an interesting direction. If the environmental maps are still 2D, one can extend the denoising network to process higher-dimensional state trajectories, as already demonstrated by Diffuser~\cite{janner2022planning}. If the environment is otherwise high-dimensional (\eg, 3D or more), the instance encoder would also need to be extended to accept higher-dimensional obstacle maps, similar to the approach used in Critical PRM~\cite{ichter2020learned}. Effective sampling around object boundaries (L9 in Alg.~\ref{alg:roadmap}) would also become non-trivial for high-dimensional environments.

\section{Conclusion}
\label{sec:conclusion}
We introduced a novel data-driven path planner named TSPDiffuser for traveling salesperson path planning problems (TSPPP) in obstacle-rich environments. Given a pair of obstacle and destination maps, TSPDiffuser samples plausible paths from a learned diffusion model and connects them to form a roadmap with a small number of essential nodes and edges. This improves the balance between solution quality and efficiency in solving TSPPPs. Our evaluation has confirmed the effectiveness of TSPDiffuser over existing classical and state-of-the-art data-driven path planners in both synthetic and real-world indoor/outdoor navigation environments.

\clearpage
\balance

\bibliographystyle{IEEEtran}
\bibliography{ref.bib}

\end{document}